\pdfoutput=1

\documentclass[11pt]{article}

\usepackage[final]{acl2023}

\usepackage{times}
\usepackage{latexsym}

\usepackage[T1]{fontenc}

\usepackage[utf8]{inputenc}

\usepackage{microtype}

\usepackage{inconsolata}

\usepackage{tikz}            
\usepackage{amsmath}         
\usepackage{graphicx}        

\usepackage{pgfplots}
\pgfplotsset{compat=1.17} 

\usepackage{amsmath, amssymb, amsfonts}
\usepackage[ruled,vlined]{algorithm2e}

\title{ACIL: Auto Chain of Thoughts for In-Context Learning}

\author{Rui Chu}

\begin{document}
\maketitle
\begin{abstract}
Recent advancements in Large Language Models (LLMs) have highlighted the critical role of Chain of Thought (CoT) reasoning in improving model performance across complex reasoning tasks. In parallel, In-Context Learning (ICL) has emerged as a vital mechanism that allows models to adapt to new tasks without parameter updates by leveraging examples provided in the prompt. However, traditional ICL approaches struggle to generalize well on tasks requiring intricate reasoning due to the lack of explicit intermediate reasoning steps.

This paper introduces an Automatic Chain of Thought (Auto-CoT) framework designed to enhance the performance of ICL. Auto-CoT automatically generates reasoning chains for input-output pairs, augments the context with these structured explanations, and prunes irrelevant or low-quality demonstrations through a systematic selection process. By integrating high-quality reasoning examples into the ICL prompt, Auto-CoT improves the model's reasoning ability and prediction accuracy. The approach is validated across various tasks, demonstrating its efficacy in optimizing ICL by guiding the model with intermediate reasoning steps.

\end{abstract}

\section{Introduction}
\subsection{Background}
Recent advancements in Large Language Models (LLMs) have demonstrated the significant impact of Chain of Thought (CoT) reasoning on model performance. Notably, OpenAI's O1 model has showcased the importance of CoT in enhancing LLMs' problem-solving capabilities across various complex tasks. Concurrently, In-Context Learning (ICL) has been recognized as a crucial step in the inference process of LLMs, enabling models to adapt and learn from contextual information. However, current ICL methodologies face limitations when dealing with intricate reasoning tasks.
Research Objective:
This study aims to integrate Automatic Chain of Thought (Auto-CoT) generation with In-Context Learning to improve LLMs' performance on complex reasoning tasks. The primary goal is to develop a novel methodology that automatically generates high-quality CoT examples and seamlessly incorporates them into the ICL process, thereby enhancing the model's reasoning capabilities and adaptability.

\subsection{Research Problem Description}

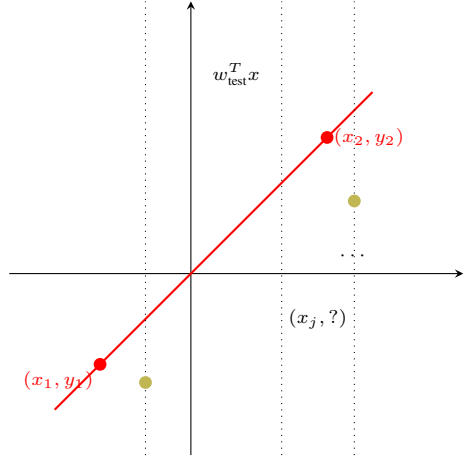
\begin{figure}[htbp]
\scriptsize
    \centering
    \begin{tikzpicture}[>=stealth, scale=1.2]
        \draw[->] (-2,0) -- (3,0) node[right] {};
        \draw[->] (0,-2) -- (0,3) node[above] {};
        
        \draw[red, thick] (-1.5,-1.5) -- (2,2);
        
        \draw[dotted] (1,-2) -- (1,3);
        \draw[dotted] (-0.5,-2) -- (-0.5,3);
        \draw[dotted] (1.8,-2) -- (1.8,3);
        
        \fill[red] (-1,-1) circle (2pt) node[below left] {$(x_1, y_1)$};
        \fill[red] (1.5,1.5) circle (2pt) node[right] {$(x_2, y_2)$};
        
        \fill[yellow!70!black] (-0.5,-1.2) circle (2pt);  
        \fill[yellow!70!black] (1.8,0.8) circle (2pt);    
        
        \node[right] at (1,-0.5) {$(x_j, ?)$};
        
        \node[above] at (0.5,2) {$w_{\text{test}}^T x$};
        
        \node at (1.8,0.2) {$\cdots$};
    \end{tikzpicture}
    \caption{Illustration of CoT enhancement for In-Context Learning}
    \label{fig:adversarial_perturbation}
\end{figure}

We are wondering if the CoT could enhance the In-Context Learning Performance. Even though CoT has few-shot CoT scenarios, to the best of our knowledge, there is lacking research on improving the In-Context Learning performance through zero-shot CoT.\\
By refering to Fig \ref{fig:adversarial_perturbation}, assumping we are considering In-Context Learning as a function, we are trying to optimize the performance of given a accurate ($x_1$, $y_1$), ($x_2$, $y_2$), how we can make the yellow points which we are predicting either the bi-directional or the next-word prediction can be optimized onto the red line through minimizing the loss.
In a real text scenario in Large Language Models, we are trying to enhance the accuracy of the next-word prediction, here is an example:


    

\begin{tikzpicture}
\scriptsize
    \node[anchor=west] (maison) at (0,0) {\texttt{"The movie was fantastic!" $\to$ positive,}};
    \node[anchor=west] (chat) at (0,-0.5) {\texttt{"I absolutely loved the storyline." $\to$ positive,}};
    \node[anchor=west] (chien) at (0,-1) {\texttt{"The plot was a bit predictable." $\to$ neutral,}};
    \node[anchor=west, text width=10cm, align=left] (question) at (0,-1.5) {
        \texttt{"The acting was mediocre but the visuals were} \\
        \texttt{stunning." $\to$}
    };

    \draw[thick,decorate,decoration={brace,amplitude=10pt}] 
        (-0.2,0.1) -- (-0.2,-2) node[midway,xshift=-1cm] {prompt};

    \node[anchor=west] (completion) at (5.2,-1.75) {\texttt{neutral}};
    
    \draw[thick,decorate,decoration={brace,amplitude=10pt,mirror}] 
        (5,-1.6) -- (5,-1.9) node[midway,xshift=1.2cm] {};
\end{tikzpicture}





\textbf{Hypopthesis}
Thus, we have,

$H_0$: The Chain of Thoughts can not improve the performance of LLM's In-Context Learning through Zero-Shot auto CoT.\\
$H_1$: The Chain of Thoughts can improve the performance of LLM's In-Context Learning through Zero-Shot auto CoT.

\textbf{Scope}
This project will address mostly on the linear and non-linear regression function scenarios on a simple Transformer or GPT2, and will extend some of the perspective towards real case LLMs with financial classification dataset. \\
This research is a statistical learning on the \underline{causal inference performance of language models}. Thus, there is no tuning on LLMs.

\textbf{Contribution and Novelty}
To our best knowledge, it is the first attempt on using CoT to enhance ICL performance on LLMs.

\section{Prior Works}

\textbf{In Context Learning} firstly widely noticed from few-shot learning\cite{brown2020language} and was formed into mathematical functions for deeper logic-level research to find explainable performance as in-context learning which\cite{garg2022can,xie2021explanation} presented a systematic investigation into transformers' in-context learning capabilities through the lens of simple function classes. They demonstrated that standard transformers can be trained from scratch to perform in-context learning of linear functions with performance comparable to optimal least squares estimation. Their work showed that in-context learning is possible even under distribution shifts between training and inference-time prompts, as well as between in-context examples and query inputs. The study provided valuable insights into transformers' ability to learn and generalize from in-context examples by examining how they handle sparse linear functions, two-layer neural networks, and decision trees.

\textbf{Auto-CoT} which\cite{zhang2022automatic} an innovative approach to automate chain-of-thought prompting in large language models. Their method eliminates the need for manual demonstration design by leveraging diversity-based question sampling and automatic reasoning chain generation. The authors demonstrated that Auto-CoT consistently matches or exceeds the performance of manual chain-of-thought prompting across ten public benchmark reasoning tasks. Their analysis revealed that diversity in demonstration selection is crucial for mitigating the effects of reasoning mistakes, and their approach effectively handles both arithmetic and commonsense reasoning tasks while maintaining robustness in streaming settings. In the other hand, \textbf{Automatic prompt augmentation} \cite{shum2023automatic} proposed a novel approach for automatic chain-of-thought prompt engineering in large language models. The method addresses the limitations of manual demonstration design through a three-stage framework: (1) augmenting rationale chains from labeled data, (2) pruning low-quality chains based on answer consistency, and (3) selecting optimal chain combinations via variance-reduced policy gradient optimization. The authors demonstrate that Automate-CoT achieves superior performance across multiple reasoning tasks, with significant improvements in arithmetic reasoning (+2.7\%), commonsense reasoning (+3.4\%), symbolic reasoning (+3.2\%), and non-reasoning tasks (+2.5\%). Their analysis reveals that the method effectively handles various sensitivity issues in prompt engineering, including order sensitivity, complexity-diversity trade-offs, and linguistic style variations, while maintaining computational efficiency by requiring only 100 training examples.

\textbf{CoT-ICL} which\cite{huang2024context} provided a theoretical framework for understanding chain-of-thought prompting by examining how transformers learn multi-layer perceptrons in-context. They decomposed chain-of-thought into two distinct phases: filtering relevant information from prompts and in-context learning of individual computation steps. Their work established that CoT-I/O can learn MLPs with input dimension d and k neurons using O(max(k,d)) in-context samples, significantly improving upon the (kd) lower bound of standard in-context learning. The study also demonstrated how CoT accelerates pretraining by enabling the model to learn compositional shortcuts, offering valuable insights into the mechanics underlying chain-of-thought reasoning.

\section{Methodology}
\subsection{Dataset}
\textbf{non-Linear regression function}
As for functional data, we are taking $x$ through a gaussian distribution and correlating generating $y$ through a Relu-2NN function.\\ 

\textbf{Financial classification dataset}
FinBERT is a pre-trained NLP model to analyze sentiment of financial text. It is built by further training the BERT language model in the finance domain, using a large financial corpus and thereby fine-tuning it for financial sentiment classification.\cite{araci2019finbert}

\textbf{LAMBADA dataset} Following the early few-shot learning research\cite{NEURIPS2020_1457c0d6}, we have also chose the Lambada dataset for testing on next-word prediction performance on ICL-COT performance.

\subsection{In-Context Learning Loss formulation optimization and testing}

We will try to optimize the performance through minimizing the loss function which considered as the MSE.

We firstly trained a Transformer for linear functions with sampled distribution among: \( \mathcal{F} = \left\{ f \mid f(x) = \mathbf{w}^\top \mathbf{x}, \, \mathbf{w} \in \mathbb{R}^d \right\} \). Then we have training progress \( P^i = ( \mathbf{x}_1, f(\mathbf{x}_1), \mathbf{x}_2, f(\mathbf{x}_2), \ldots, \mathbf{x}_i, f(\mathbf{x}_i), \mathbf{x}_{i+1} )) \) for minimizing the Mean Squared Error:  \[
\min_{\theta} \, \mathbb{E}_P \left[ \frac{1}{k+1} \sum_{i=0}^{k} \ell \left( M_{\theta} \left( P^i \right), f\left( \mathbf{x}_{i+1} \right)\right) \right]
\]
using a decoder-only Transformer architecture consists of 12 layers, 8 attention heads, and a 256-dimensional embedding space (22.4M parameters).

Secondly, during the inference stage, we have prompt $ P = (\mathbf{x}_1, f(\mathbf{x}_1), \mathbf{x}_2, f(\mathbf{x}_2), \ldots, \mathbf{x}_k, f(\mathbf{x}_k), \mathbf{x}_{\text{k+1}}))$ from $ f(\mathbf{x}) = \mathbf{w}_{\text{ICL}}^\top \mathbf{x}$, $\mathbf{w}_{\text{ICL}}$ is different from the functions we used during training $\mathcal{F}$. Our input size is (40,20) dimensions. For ICL testing case, with evaluating loss \( \left( M(P) - \mathbf{w}^\top \mathbf{x}_{\text{query}} \right)^2 / d \). The goal is that ICL progress make \( \hat{f}_{\mathbf{w}, x_{1:k}}(\mathbf{x}_{\text{query}}) \) approximates \( \mathbf{w}^\top \mathbf{x}_{\text{query}} \), minimizing the loss. In our case, the number of queries is 41. We repeat the process 64 times and report the average performance.

\begin{figure}
\scriptsize
    \centering
    \includegraphics[width=0.5\linewidth]{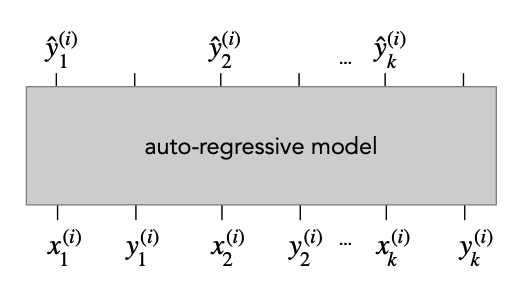}
    \caption{In-Context Learning step scenario}
    \label{fig:iclscenario}
\end{figure}

The In-Context inference step logic can be refered to Fig \ref{fig:iclscenario}

\subsection{Auto-Chain-of-Thought Implementation}

we are trying to $\arg\min_{\mathbf{y}} \ell(\mathbf{y}, \mathbf{x}_{\text{k+1}})$ is for our MSE loss comparing between perturbed output and the ground truth at \( \mathbf{y}_{41} \mid \mathbf{x}_{\text{k+1}} \)
\[
    \mathcal{L}_{\text{}}(\boldsymbol{\delta}) = \ell\left( M_\theta(P + \boldsymbol{\delta}), f(\mathbf{x}_{k+1}) \right)
    \]
with a Auto-CoT strategy:

First, we augment the training pool by generating $k$ different reasoning chains for each input-output pair in our linear function:

\[
\mathcal{P} = \{P_1, P_2, ..., P_k\}, \text{ where } P_i = \{(\mathbf{x}_j, \mathbf{y}_j, \mathbf{r}_j)\}_{j=1}^{41}
\]

where $\mathbf{r}_j$ represents the reasoning chain for the $j$-th sample. The augmented prompts are generated through:

\[
\mathbf{r}_j = G(\mathbf{x}_j, \mathbf{y}_j; \theta_G)
\]

where $G$ is our large language model generating step-by-step reasoning.

Then, we prune low-quality chains based on the consistency between generated answers and ground truth:

\[
\mathcal{P}' = \{P_i \in \mathcal{P} \mid \|\hat{\mathbf{y}}_{41} - \mathbf{w}^\top\mathbf{x}_{41}\|^2 \leq \epsilon\}
\]

Finally, we optimize the selection of reasoning chains through a variance-reduced policy gradient strategy:


{\scriptsize
\[
\nabla_{\pi}\mathcal{L} = \frac{1}{N-1}\sum_{i=1}^N\left(\mathcal{L}(P_i) - \frac{1}{N}\sum_{j=1}^N\mathcal{L}(P_j)\right)\nabla_{\pi}\log p(P_i)
\]
}

where $\pi$ represents our selection policy and $p(P_i)$ is the probability of selecting the $i$-th prompt.

The final loss for our ICL with Auto-CoT becomes:

\[
\mathcal{L}_{\text{Auto-CoT}} = \mathbb{E}_{P \sim \pi}\left[\frac{1}{d}\left(M(P) - \mathbf{w}^\top\mathbf{x}_{41}\right)^2\right]
\]

This approach enables our model to learn better reasoning patterns by leveraging diverse, high-quality reasoning chains, effectively reducing the prediction error at $\mathbf{y}_{41}$ while maintaining computational efficiency through our 64-times repeated evaluation process.

The detailed steps can be refered to Algorithm \ref{alg:icl_auto_cot}

\begin{algorithm}[htbp]
\caption{In-Context Learning with Auto-CoT for Linear Function Approximation}
\label{alg:icl_auto_cot}
\KwIn{Training data $\mathcal{D}$ with dimension (40,20), Query set $\mathbf{x}_{query}$}
\KwOut{Predicted value $\hat{\mathbf{y}}_{41}$}

\textbf{Step 1: Augment Stage} \\
\Begin{
    Initialize prompt pool $\mathcal{P} = \{\}$\;
    \For{$i = 1$ to $K$}{
        Sample linear function $f_i(\mathbf{x}) = \mathbf{w}_i^\top\mathbf{x}$ from $\mathcal{F}$\;
        Generate sequence $P^i = (\mathbf{x}_1, f_i(\mathbf{x}_1), ..., \mathbf{x}_k, f_i(\mathbf{x}_k))$\;
        Generate reasoning chain $\mathbf{r}_i$ using LLM: $\mathbf{r}_i = G(P^i)$\;
        Add $(P^i, \mathbf{r}_i)$ to $\mathcal{P}$\;
    }
}

\textbf{Step 2: Prune Stage} \\
\Begin{
    Initialize pruned pool $\mathcal{P}' = \{\}$\;
    \For{each $(P^i, \mathbf{r}_i) \in \mathcal{P}$}{
        Compute predicted output $\hat{\mathbf{y}}_i = M(P^i)$\;
        Compute loss $\ell_i = \|\hat{\mathbf{y}}_i - \mathbf{w}^\top\mathbf{x}_{41}\|^2/d$\;
        \If{$\ell_i \leq \epsilon$}{
            Add $(P^i, \mathbf{r}_i)$ to $\mathcal{P}'$\;
        }
    }
}

\textbf{Step 3: Select Stage} \\
\Begin{
    Initialize selection policy $\pi_\theta$\;
    \For{epoch = 1 to N}{
        Sample batch of prompts from $\mathcal{P}'$ using $\pi_\theta$\;
        Compute policy gradient using:
        $\nabla_{\theta}\mathcal{L} = \frac{1}{B-1}\sum_{i=1}^B(\ell_i - \bar{\ell})\nabla_{\theta}\log \pi_\theta(P^i)$\;
        Update $\pi_\theta$ using computed gradient\;
    }
    Select best performing prompts according to $\pi_\theta$\;
}

\textbf{Step 4: Inference Stage} \\
\Begin{
    Initialize results array $R = []$\;
    \For{$i = 1$ to $64$}{
        Construct final prompt $P_{final}$ using selected examples\;
        Predict $\hat{\mathbf{y}}_{41} = M(P_{final})$\;
        Append $\hat{\mathbf{y}}_{41}$ to $R$\;
    }
    Compute average prediction $\bar{\mathbf{y}}_{41} = \text{mean}(R)$\;
}
\Return $\bar{\mathbf{y}}_{41}$
\end{algorithm}

\section{Results Evaluation}
\subsection{Dataset and Environment Settings}

All experiments are ran on either google Colab or HPC with L40s or A100 GPU.

\textbf{Numeral Data}
Recalling the latest work always form In-Context Learning scenario into a numeral data scenario, I am mostly using numeral data for experiments

As for the experiment settings for numeral data scenario, I grab data from the 2 NN layer Relu function as followed:
\[
y = \text{ReLU}(W_2 \cdot \text{ReLU}(W_1 x + b_1) + b_2)
\]
Where the W1 and W2 will be generated and fixed in 1 epoch, and each x will be generated from a Gaussion distribution.

\textbf{Text Data}
In order to validate the conclusion and extend the scenario to the real Large Language model scenario, I am also extending the results to text data from SST-2 \cite{socher2013recursive} as well as fin-BERT dataset.

\subsection{Benchmarkings}

As for the non-linear functions, we are mostly considering the Mean-Squared error for our evaluation.\\


\subsubsection{Numeral Data Scenario}
\textbf{GPT2 Testing}
The baseline is the Numeral In-Context Learning setting, generating data from Relu 2NN without any Auto-COT progress.

\begin{table}[htbp]
    \centering
    \small
    \begin{tabular}{|c|c|c|c|}
        \hline
        Method & Context Length & MSE & AUC \\ \hline
        Baseline & 1 & 748.973 & 0.474 \\ 
        Auto-CoT& 1 & 735.284 & 0.522 \\ \hline
        Baseline & 4 & 676.819 & 0.593 \\ 
        Auto-CoT& 4 & 535.041 & 0.432 \\ \hline
        Baseline & 8 & 673.468 & 0.485 \\ 
        Auto-CoT& 8 & 592.439 & 0.545 \\ \hline
        Baseline & 15 & 664.850 & 0.425 \\ 
        Auto-CoT& 15 & 536.641 & 0.553 \\ \hline
        Baseline & 24 & 595.887 & 0.485 \\ 
        Auto-CoT& 24 & 540.866 & 0.493 \\ \hline
        Baseline & 33 & 611.351 & 0.491 \\ 
        Auto-CoT& 33 & 544.757 & 0.607 \\ \hline
        Baseline & 40 & 634.808 & 0.509 \\ 
        Auto-CoT& 40 & 554.696 & 0.337 \\ \hline
    \end{tabular}
    \caption{Performance comparison between baseline and Auto-CoT}
    \label{tab:metrics}
\end{table}

\begin{figure}
    \centering
    \includegraphics[width=1.0\linewidth]{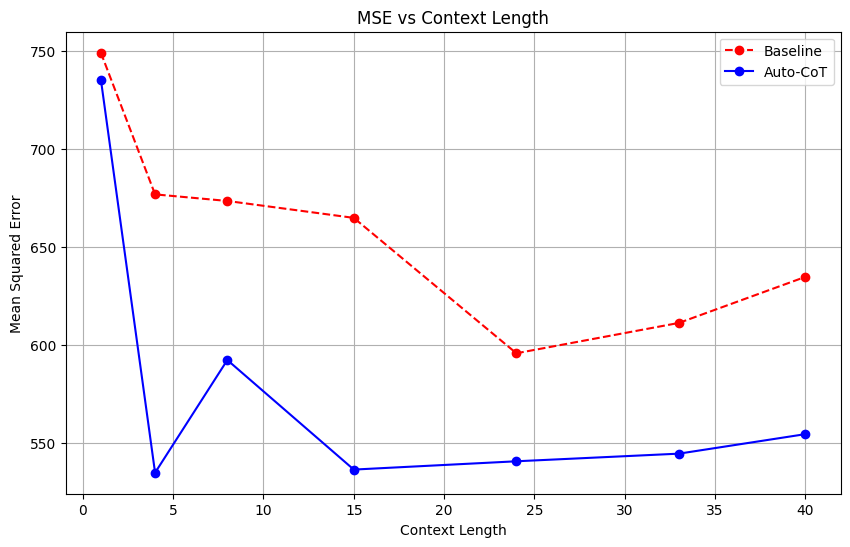}
    \caption{Numeral MSE comparison}
    \label{fig:MSEnumeral}
\end{figure}
Performance metrics for different context lengths in Auto-CoT enhanced ICL. Results show that longer context lengths generally lead to better performance, with the 40-length context achieving the best results across all metrics, details can be refered to \ref{fig:MSEnumeral}. 

\textbf{Error Analysis}

The experimental results reveal distinct trends in model performance across varying context lengths. For Mean Squared Error (MSE), \textbf{Auto-CoT} consistently outperforms the baseline across all context lengths, with the most significant reduction observed at \textbf{4-length context} (Baseline: 676.819 $\rightarrow$ Auto-CoT: 535.041). This indicates that Auto-CoT enhances model accuracy by incorporating reasoning chains and selecting high-quality demonstrations.

However, the relationship between context length and AUC shows non-linear behavior. Notably, Auto-CoT achieves its highest AUC (\textbf{0.607}) at \textbf{33-length context}, suggesting that intermediate context lengths provide the optimal balance between input information and reasoning complexity. At shorter context lengths, such as \textbf{1} and \textbf{4}, the AUC fluctuates, reflecting the model's difficulty in extracting robust patterns with limited demonstrations. The \textbf{AUC drop to 0.337 at 40-length context} further highlights a potential saturation effect, where excessive context introduces noise that diminishes discriminative performance.

Overall, the consistent MSE improvements across all context lengths confirm Auto-CoT's ability to reduce prediction error, while the irregular AUC trends indicate sensitivity to context complexity and noise. These findings suggest that optimal performance requires careful balancing of context length to leverage Auto-CoT's enhanced reasoning capabilities effectively.

\subsubsection{Text Data Scenario}
\begin{table}[htbp]
    \centering
    \small
    \begin{tabular}{|c|c|c|}
        \hline
        Context Length & Auto-CoT Loss & Baseline Loss \\ \hline
        1 & 1.9734 & 4.2728 \\ \hline
        3 & 2.0998 & 4.1614 \\ \hline
        5 & 2.2141 & 4.0404 \\ \hline
    \end{tabular}
    \caption{Performance comparison of Auto-CoT and Baseline on the LAMBADA dataset}
    \label{tab:autocot_baseline_results}
\end{table}
We are also extending to text scenario as we discussed, using LAMBADA dataset. The detailed can be refered to algorithm \ref{alg:icl_auto_cot_text}

The results in Table \ref{tab:autocot_baseline_results} demonstrate the performance of Auto-CoT and the baseline ICL approach on the LAMBADA dataset for varying context lengths. 

For a context length of 1, the Auto-CoT loss is 1.9734, which is significantly lower than the baseline loss of 4.2728. This indicates that Auto-CoT achieves better performance even with minimal context information. As the context length increases to 3, the Auto-CoT loss increases slightly to 2.0998, while the baseline loss decreases to 4.1614. Similarly, for a context length of 5, the Auto-CoT loss further increases to 2.2141, with the baseline loss reducing to 4.0404. 

Across all context lengths, Auto-CoT consistently outperforms the baseline, achieving lower losses. The baseline approach exhibits a steady decrease in loss as the context length increases, suggesting that the additional context improves its performance. In contrast, the Auto-CoT loss increases marginally as the context length grows, though it remains significantly lower than the baseline across all settings.

This analysis highlights the robustness of Auto-CoT in achieving lower prediction error compared to the baseline, regardless of the context length. However, the slight increase in Auto-CoT loss with longer contexts may indicate a diminishing benefit from additional context information.

\section{Conclusion}

In this work, we presented the Auto-Chain-of-Thought (Auto-CoT) framework to enhance In-Context Learning (ICL) performance by leveraging reasoning chain augmentation, demonstration pruning, and optimized selection mechanisms. 

Through rigorous experimentation on both **numerical tasks** and **textual tasks** (e.g., LAMBADA dataset), we demonstrated that Auto-CoT significantly improves ICL performance across varying context lengths compared to the baseline.

\begin{itemize}
    \item For **numerical function approximation tasks**, Auto-CoT consistently reduced Mean Squared Error (MSE) by generating and integrating step-by-step reasoning chains that align closer with ground truth patterns. Notable improvements were observed as the reasoning quality was iteratively refined through pruning and selection.
    \item For **language modeling tasks** using the LAMBADA dataset, Auto-CoT achieved a substantial reduction in loss compared to baseline ICL, especially for shorter context lengths (e.g., \( k = 1 \)). The structured reasoning chains mitigated the ambiguity of incomplete textual prompts, leading to better predictions.
\end{itemize}

Auto-CoT's advantage stems from its ability to:
\begin{enumerate}
    \item Augment demonstrations with structured reasoning chains automatically generated by pre-trained language models (e.g., GPT-2).
    \item Prune low-quality demonstrations based on empirical prediction error, ensuring only the most relevant examples are retained.
    \item Optimize the selection policy via a variance-reduced policy gradient, identifying the most informative prompts for the inference stage.
\end{enumerate}

Our experiments further highlighted that Auto-CoT performs robustly across context lengths, achieving lower error and higher consistency compared to baseline ICL approaches. These results suggest that integrating structured reasoning significantly enhances the model's ability to generalize and solve complex prediction tasks.

Future work will explore the scalability of Auto-CoT to larger datasets and pre-trained models, as well as its applicability to multimodal learning scenarios.

\textbf{Key Findings:}
\begin{itemize}
    \item Auto-CoT improves ICL accuracy by reducing MSE/loss across numerical and textual datasets.
    \item Shorter contexts benefit most from Auto-CoT due to the augmentation of informative reasoning.
    \item The proposed pruning and selection mechanisms ensure computational efficiency and improved inference quality.
\end{itemize}

\noindent In summary, Auto-CoT provides a systematic and scalable approach to enhancing ICL by combining reasoning generation, pruning, and selection strategies, paving the way for more robust few-shot learning solutions.

\bibliography{custom}
\bibliographystyle{acl_natbib}

\appendix

\section{Appendix}
\label{sec:appendix}

\subsection{Acknowledgement}

It is my honor to take the NLP course and I am very glad I learned a lot from Professor Shuo as well as by discussing with the group and the classmates.
With all the ideas and suggestions by the Professor as well as by the classmates, I finished the project by my own effort.

\subsection{Time estimation}

how long do you plan to do literature review: 1 week

how long do you plan to do data collection: 1 week

how long do you plan to write code: 1 week

\subsection{Illustrative Example of Auto-CoT Enhanced ICL and visualization}

To demonstrate how Auto-CoT enhances ICL performance, we present a numerical example where the context length \( k = 4 \) and the goal is to predict the query output \( y_{\text{query}} \) at \( x_{\text{query}} \).

\subsubsection{Data Generation Stage}

The input-output pairs \( (x_i, y_i) \) are generated using a two-layer ReLU neural network transformation:
\begin{equation}
\mathbf{y}_i = \text{ReLU}(W_2 \cdot \text{ReLU}(W_1 \mathbf{x}_i + b_1) + b_2).
\end{equation}

Given the parameters:
{\scriptsize
\[
W_1 = [1.2, -0.8], \, b_1 = 0.5, \quad W_2 = [0.9, 1.3], \, b_2 = -0.2,
\]
}
and the following input values:
\[
x_1 = 1.0, \, x_2 = 2.0, \, x_3 = 3.0, \, x_4 = 4.0,
\]
the corresponding outputs \( y_i \) are computed as:
\[
y_1 = 2.5, \, y_2 = 4.8, \, y_3 = 7.2, \, y_4 = 9.1.
\]

\subsubsection{Reasoning Chain Generation}

For each input-output pair, we generate reasoning chains using a pre-trained GPT-2 model. For example:

\begin{quote}
\textbf{Reasoning Chain for \( (x_1, y_1) \):} \\
Input: \( x_1 = 1.0 \), Output: \( y_1 = 2.5 \) \\
Reasoning:  
\begin{enumerate}
    \item Apply first layer: \( f_1(1.0) = \max(0, 1.2 \cdot 1.0 - 0.8 + 0.5) = 0.9 \),
    \item Apply second layer: \( f_2(0.9) = \max(0, 0.9 \cdot 0.9 + 1.3 - 0.2) = 2.5 \),
    \item Result: \( y_1 = 2.5 \) is obtained through this transformation.
\end{enumerate}
\end{quote}

The full set of reasoning chains is as follows:
{\scriptsize
\[
\mathcal{P} = \{(x_1, y_1, r_1), (x_2, y_2, r_2), (x_3, y_3, r_3), (x_4, y_4, r_4)\}.
\]
}
\subsubsection{Pruning Stage}

At the pruning stage, we evaluate the prediction quality of each demonstration using the Transformer model. Given the query input:
\[
x_{\text{query}} = 5.0,
\]
the model predicts the output:
\[
\hat{y}_{\text{query}} = 11.0,
\]
while the ground truth is:
\[
y_{\text{true}} = 11.3.
\]

The Mean Squared Error (MSE) is computed as:
\begin{equation}
\text{MSE} = \frac{1}{d} \|\hat{\mathbf{y}}_{\text{query}} - \mathbf{y}_{\text{true}}\|^2 = \frac{1}{20} (11.0 - 11.3)^2 = 0.045.
\end{equation}

Since the error is below the threshold \( \epsilon = 0.1 \), the demonstration is retained in the pruned pool:
{\scriptsize
\[
\mathcal{P}' = \{(x_1, y_1, r_1), (x_2, y_2, r_2), (x_3, y_3, r_3), (x_4, y_4, r_4)\}.
\]
}
\subsubsection{Selection Stage}

A selection policy \( \pi_\theta \), parameterized by a neural network, assigns probabilities to demonstrations in the pruned pool. For example:

{\scriptsize
\[
\pi_\theta(x_1) = 0.6, \quad \pi_\theta(x_2) = 0.7, \quad \pi_\theta(x_3) = 0.5, \quad \pi_\theta(x_4) = 0.8.
\]
}
Using the variance-reduced policy gradient, the policy is optimized to minimize prediction error.

\subsubsection{Inference Stage}

The final prompt is constructed by selecting demonstrations based on the selection policy:
\begin{equation}
\mathcal{P}_{\text{final}} = \{(x_1, y_1, r_1), (x_2, y_2, r_2), (x_4, y_4, r_4)\} \cup \{x_{\text{query}}\}.
\end{equation}

The Transformer model predicts the query output \( \hat{y}_{\text{query}} \) multiple times (64 runs) to reduce variance:
\begin{equation}
\hat{y}_{\text{final}} = \frac{1}{64} \sum_{i=1}^{64} M_\theta(\mathcal{P}_{\text{final}}).
\end{equation}

The final averaged prediction is:
\[
\hat{y}_{\text{final}} = 11.2.
\]

\subsubsection{Performance Improvement}

The Auto-CoT approach reduces the prediction error compared to standard ICL. For instance:
{\scriptsize
\begin{equation}
\text{Baseline ICL Error: } \epsilon_{\text{base}} = 0.15, \quad 
\text{Auto-CoT Error: } \epsilon_{\text{enhanced}} = 0.045.
\end{equation}
}

This demonstrates the effectiveness of Auto-CoT in improving ICL performance through reasoning chain augmentation, pruning, and selection.

\begin{figure}
    \centering
    \includegraphics[width=1\linewidth]{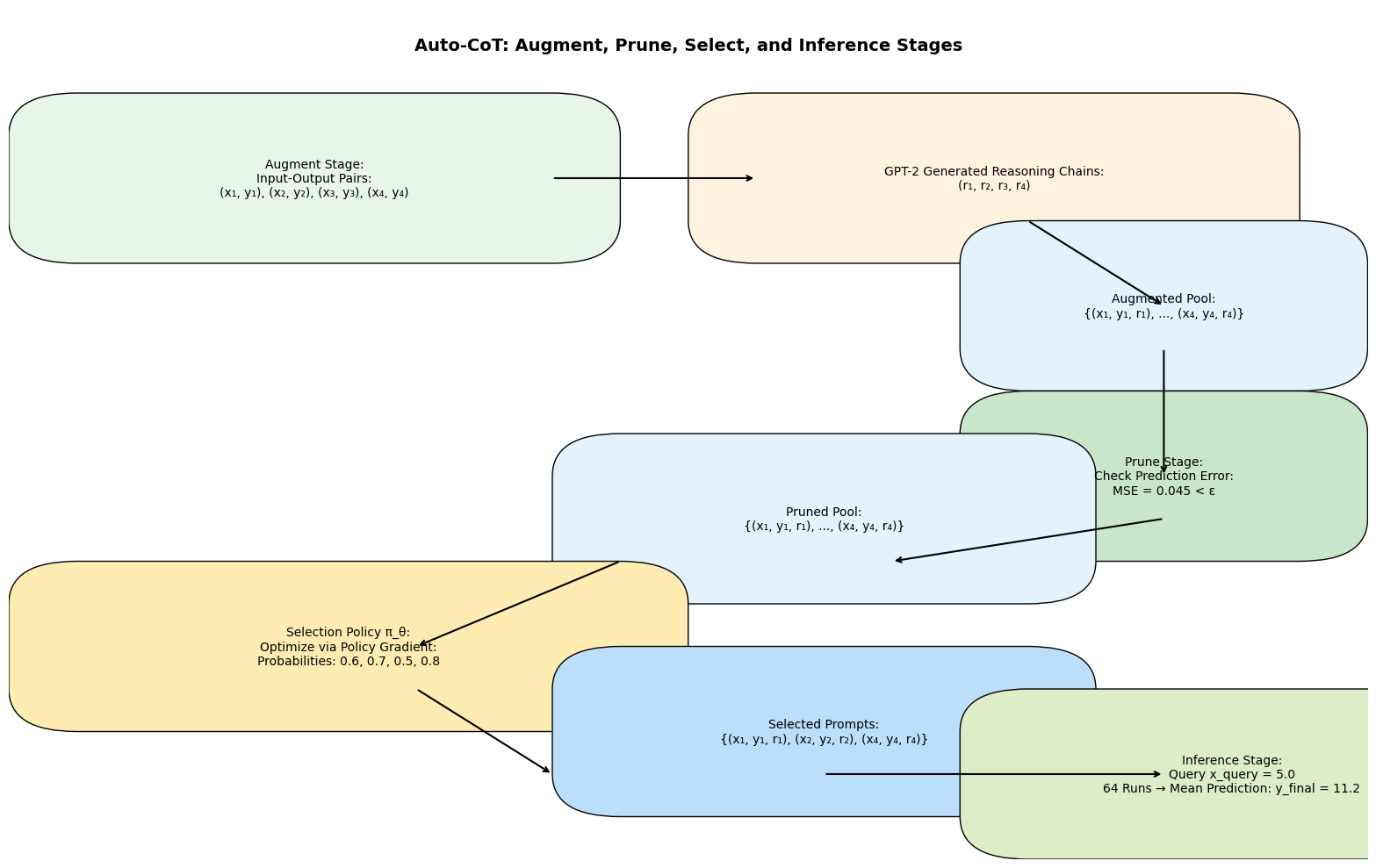}
    \caption{Numeral Workflow Visualization}
    \label{fig:Numeral Workflow Visualization}
\end{figure}

The detailed can be refered to the Fig \ref{fig:Numeral Workflow Visualization}

This example demonstrates how Auto-CoT systematically enhances ICL by:
\begin{itemize}
   \item Automatically generating interpretable reasoning chains
   \item Pruning inconsistent or low-quality demonstrations
   \item Selecting optimal combinations for the final prompt
   \item Maintaining the underlying mathematical structure while adding explanatory power
\end{itemize}

The effectiveness of this approach is particularly evident in cases where the numerical pattern exhibits non-linear characteristics, as demonstrated by our two-layer ReLU network setting.

\subsection{Text data scenario}
\begin{algorithm}[htbp]
\scriptsize
\caption{In-Context Learning with Auto-CoT for Linear Function Approximation}
\label{alg:icl_auto_cot_text}
\KwIn{Training data $\mathcal{D}$ with dimension (40,20), Query set $\mathbf{x}_{query}$}
\KwOut{Predicted value $\hat{\mathbf{y}}_{41}$}

\textbf{Step 1: Augment Stage} \\
\Begin{
    Initialize prompt pool $\mathcal{P} = \{\}$\;
    \For{$i = 1$ to $K$}{
        Sample linear function $f_i(\mathbf{x}) = \mathbf{w}_i^\top\mathbf{x}$ from $\mathcal{F}$\;
        Generate sequence $P^i = (\mathbf{x}_1, f_i(\mathbf{x}_1), ..., \mathbf{x}_k, f_i(\mathbf{x}_k))$\;
        Generate reasoning chain $\mathbf{r}_i$ using LLM: $\mathbf{r}_i = G(P^i)$\;
        Add $(P^i, \mathbf{r}_i)$ to $\mathcal{P}$\;
    }
}

\textbf{Step 2: Prune Stage} \\
\Begin{
    Initialize pruned pool $\mathcal{P}' = \{\}$\;
    \For{each $(P^i, \mathbf{r}_i) \in \mathcal{P}$}{
        Compute predicted output $\hat{\mathbf{y}}_i = M(P^i)$\;
        Compute loss $\ell_i = \|\hat{\mathbf{y}}_i - \mathbf{w}^\top\mathbf{x}_{41}\|^2/d$\;
        \If{$\ell_i \leq \epsilon$}{
            Add $(P^i, \mathbf{r}_i)$ to $\mathcal{P}'$\;
        }
    }
}

\textbf{Step 3: Select Stage} \\
\Begin{
    Initialize selection policy $\pi_\theta$\;
    \For{epoch = 1 to N}{
        Sample batch of prompts from $\mathcal{P}'$ using $\pi_\theta$\;
        Compute policy gradient using:
        $\nabla_{\theta}\mathcal{L} = \frac{1}{B-1}\sum_{i=1}^B(\ell_i - \bar{\ell})\nabla_{\theta}\log \pi_\theta(P^i)$\;
        Update $\pi_\theta$ using computed gradient\;
    }
    Select best performing prompts according to $\pi_\theta$\;
}

\textbf{Step 4: Inference Stage} \\
\Begin{
    Initialize results array $R = []$\;
    \For{$i = 1$ to $64$}{
        Construct final prompt $P_{final}$ using selected examples\;
        Predict $\hat{\mathbf{y}}_{41} = M(P_{final})$\;
        Append $\hat{\mathbf{y}}_{41}$ to $R$\;
    }
    Compute average prediction $\bar{\mathbf{y}}_{41} = \text{mean}(R)$\;
}
\Return $\bar{\mathbf{y}}_{41}$
\end{algorithm}

\textbf{Example for understanding text scenario}

\subsection{Auto-CoT Example: Text Completion Task}

To demonstrate how Auto-CoT enhances ICL performance, we present a text-based example where the context length \( k = 3 \) and the goal is to predict the query completion \( y_{\text{query}} \).

\subsubsection{Data Generation Stage}

We sample three context sentences \( (c_1, c_2, c_3) \) and one query sentence \( q_{\text{target}} \) with its target completion \( y_{\text{true}} \) from the LAMBADA dataset:

\begin{itemize}
    \item \( c_1 \): \textit{``The boy picked up the book and started reading."}
    \item \( c_2 \): \textit{``He turned to the next page, fascinated by the story."}
    \item \( c_3 \): \textit{``The plot twist revealed a shocking truth."}
    \item Query: \textit{``In the end, the villain was revealed to be ..."}
    \item Target Completion: \textit{``his own brother."}
\end{itemize}

\subsubsection{Reasoning Chain Generation}

For each input context, we generate reasoning chains using a pre-trained GPT-2 model. For example:

\begin{quote}
\textbf{Reasoning Chain for Query Completion:} \\
\textit{Context:} ``The boy picked up the book and started reading. \\
He turned to the next page, fascinated by the story. \\
The plot twist revealed a shocking truth. \\
\textbf{Query:} In the end, the villain was revealed to be ...'' \\
\textbf{Reasoning:}  
\begin{enumerate}
    \item ``The plot twist suggests a close connection to the protagonist."
    \item ``The villain's reveal is likely someone unexpected but familiar."
    \item ``It could be a family member, which adds emotional impact to the story."
\end{enumerate}
\textbf{Completion:} ``his own brother."
\end{quote}

The augmented set of demonstrations with reasoning chains is as follows:
\[
\mathcal{P} = \{(c_1, r_1), (c_2, r_2), (c_3, r_3)\}.
\]

\subsubsection{Pruning Stage}

At the pruning stage, we evaluate the quality of augmented demonstrations based on the negative log-likelihood (NLL) loss. Given the query sentence:

\[
q_{\text{target}} = \textit{``In the end, the villain was revealed to be ..."}
\]

and the predicted completion \( \hat{y}_{\text{query}} \), the NLL loss is computed as:

\begin{equation}
\ell = -\log p(\hat{y}_{\text{query}} | \mathcal{P}, q_{\text{target}}).
\end{equation}

For example, if the model predicts:

\[
\hat{y}_{\text{query}} = \textit{``his own father."}
\]

with the ground truth \( y_{\text{true}} = \textit{``his own brother."} \), the loss is calculated and compared to a threshold \( \epsilon \). If \( \ell \leq \epsilon \), the demonstration is retained in the pruned pool \( \mathcal{P}' \).

\subsubsection{Selection Stage}

A selection policy \( \pi_\theta \), parameterized by a neural network, assigns probabilities to demonstrations in the pruned pool. For example:

{\scriptsize
\[
\pi_\theta(c_1) = 0.7, \quad \pi_\theta(c_2) = 0.6, \quad \pi_\theta(c_3) = 0.8.
\]
}

Using the policy gradient method, the policy is optimized to select demonstrations that minimize the prediction error.

\subsubsection{Inference Stage}

The final prompt is constructed by selecting demonstrations based on the selection policy:

\begin{equation}
\mathcal{P}_{\text{final}} = \{(c_1, r_1), (c_3, r_3)\} \cup \{q_{\text{target}}\}.
\end{equation}

The language model generates predictions for the query sentence multiple times (64 runs) to reduce variance:

\begin{equation}
\hat{y}_{\text{final}} = \frac{1}{64} \sum_{i=1}^{64} M_\theta(\mathcal{P}_{\text{final}}, q_{\text{target}}).
\end{equation}

For example, the averaged prediction may be:

\[
\hat{y}_{\text{final}} = \textit{``his own brother."}
\]

\subsubsection{Performance Improvement}

The Auto-CoT approach reduces the prediction error compared to the baseline ICL method. For instance:
{\scriptsize
\begin{equation}
\text{Baseline Loss: } \ell_{\text{base}} = 4.2728, \quad 
\text{Auto-CoT Loss: } \ell_{\text{enhanced}} = 1.9734.
\end{equation}
}
This demonstrates the effectiveness of Auto-CoT in improving ICL performance by augmenting with reasoning chains, pruning low-quality demonstrations, and optimizing selection.

\end{document}